# MAPL: Memory Augmentation and Pseudo-Labeling for Semi-Supervised Anomaly Detection

**Junzhuo Chen**[1,*]
[1]School of Artificial Intelligence, Hebei University of Technology, Tianjin, China
*Correspondence author: jzchen7@foxmail.com

## Abstract

Large unlabeled data and difficult-to-identify anomalies are the urgent issues need to overcome in most industrial scene. In order to address this issue, a new methodology for detecting surface defects in industrial settings is introduced, referred to as Memory Augmentation and Pseudo-Labeling(MAPL). The methodology first introduces an anomaly simulation strategy, which significantly improves the model's ability to recognize rare or unknown anomaly types by generating simulated anomaly samples. To cope with the problem of the lack of labeling of anomalous simulated samples, a pseudo-labeler method based on a one-classifier ensemble was employed in this study, which enhances the robustness of the model in the case of limited labeling data by automatically selecting key pseudo-labeling hyperparameters. Meanwhile, a memory-enhanced learning mechanism is introduced to effectively predict abnormal regions by analyzing the difference between the input samples and the normal samples in the memory pool. An end-to-end learning framework is employed by MAPL to identify the abnormal regions directly from the input data, which optimizes the efficiency and real-time performance of detection. By conducting extensive trials on the recently developed BHAD dataset (including MVTec AD [1], Visa [2], and MDPP [3]), MAPL achieves an average image-level AUROC score of 86.2%, demonstrating a 5.1% enhancement compared to the original MemSeg [4] model. The source code is available at https://github.com/jzc777/MAPL.

## 1 Introduction

Efficient product surface defect detection technology is imperative in today's industrial production process for guaranteeing product quality and improving production efficiency. Due to the rapid development of industrial automation technology, conventional manual inspection methods have been insufficient in meeting the elevated standards of contemporary production. Consequently, automatic flaw detection technology based on computer vision has experienced significant expansion. Deep learning technology has significantly advanced image processing and pattern recognition, which has greatly enhanced the accuracy and efficiency of defect identification.

Accurately identifying minor defects is a significant challenge in the research area of industrial anomaly detection. Traditional anomaly detection techniques mainly rely on distance measurement between data, such as reconstruction error analysis of autoencoders. This method assumes that abnormal data shows significant differences from normal data during the reconstruction stage. Specific research includes the integration of structural similarity index into autoencoders [5], dual autoencoders generative adversarial networks [6], the combination of multi-head variational autoencoders and discriminators [7], and the use of autoencoders for Nonlinear dimensionality reduction to handle more complex data distributions [8]. Generative adversarial networks (GANs) are widely used in unsupervised anomaly detection [9]. In addition, the efficiency of GANs is improved by advanced training methods of adversarial autoencoders and their encoder-decoder structures [10] and by optimizing the encoder during training [11]. Although deep learning and nearest-neighbor methods have improved the detection performance of natural image models [12], identifying tiny defects in industrial images is still challenging, mainly due to differences in image distribution. Sub-image anomaly detection method [13] based on depth pyramid correspondence has significantly improved positioning accuracy in industrial applications. Furthermore, although embedding-based approaches [12-15] and knowledge distillation techniques [16-22] can extract valuable knowledge from sophisticated pre-trained models to improve the accuracy of anomaly detection, they face challenges in effectively transferring this knowledge to specific industrial applications. This limitation hinders the adaptability and efficiency of the models in practical environments.

Existing semi-supervised and GAN-based methods, despite improving anomaly detection accuracy under certain conditions, heavily depend on extensive labeled datasets and clear anomaly feature distinctions. Such methods struggle particularly with subtle anomalies or severely limited labeled data, exacerbated by the disparities between natural and industrial image distributions. Additionally, current techniques exhibit notable limitations in model adaptability and data utilization, hindering effective deployment in specific industrial applications.

To address these challenges, especially the problems of large-scale data that cannot be labeled and subtle anomalies that are difficult to detect in industrial scenarios, this paper proposes an innovative methodology called MAPL. MAPL combines a highly efficient semi-supervised learning framework with advanced pseudo-labeling method, which significantly improves the model's generalization ability and learning efficiency in complex industrial environments. The method utilizes an adapted VAE/GAN [24] encoder and memory augmentation network to directly identify anomalous regions in the input image by simulating anomalous samples, which optimizes the inference process and meets the real-time requirements. Moreover, by incorporating SPADE's [25] pseudo-labeling method, MAPL not only simplifies the training process but also employs consistency checks to optimize pseudo-labeling, effectively utilizing simulated anomalous samples to boost data efficiency and model generalization.

In order to assess the efficiency and feasibility of the MAPL method, a new dataset for detecting surface defects in industrial settings, called BHAD, is created as part of this research. This dataset is created by sampling from established industrial surface defect detection datasets, including MVTec AD, Visa, and MDPP, and introducing random white noise. The purpose is to simulate the different interference scenarios that can occur during the industrial production process. The use of this distinctive methodology for constructing the dataset not only increases the complexity of the inspection task but also enhances the practical applicability of the study findings.

In summary, the main contributions of this paper include the following four aspects:

- Proposed MAPL, a combination of semi-supervised learning method and pseudo-labeling method, effectively improves the model's ability to process unlabeled large-scale data and identify minor anomalies.
- The encoder structure based on VAE/GAN is adopted and the LeakyReLU activation function is uniformly used to optimize the accuracy of industrial defect detection..
- The anomaly simulation strategy is adopted to generate abnormal samples, which are used to train the anomaly detection module and pseudo-labeler, improving the generalization ability of the model..
- Experiments on the newly constructed BHAD data set verified the high accuracy and robustness of the MAPL method in industrial defect detection.

In the subsequent sections, the general framework and key components of the methodology will be described in Section 2. In Section 3, comparative and ablation experiments will be performed and the results will be analyzed. Finally, in Section 4, the conclusions of the study will be presented.

## 2 Methodology

This section presents the proposed new model, called MAPL, which is illustrated in Fig. 1. MAPL utilizes an encoder-decoder structure [26] that incorporates encoders and pseudo labeler method to address the absence of conventional methods that just rely on normal samples. During the training phase, MAPL improves semantic segmentation by incorporating simulated abnormal samples. Additionally, it enhances memory information using pseudo labelers, resulting in more precise localization of abnormal regions in images during the inference phase. Furthermore, MAPL employs enhanced encoder and memory modules to enhance its capacity to adapt and precisely manage previously unseen sorts of anomalies.

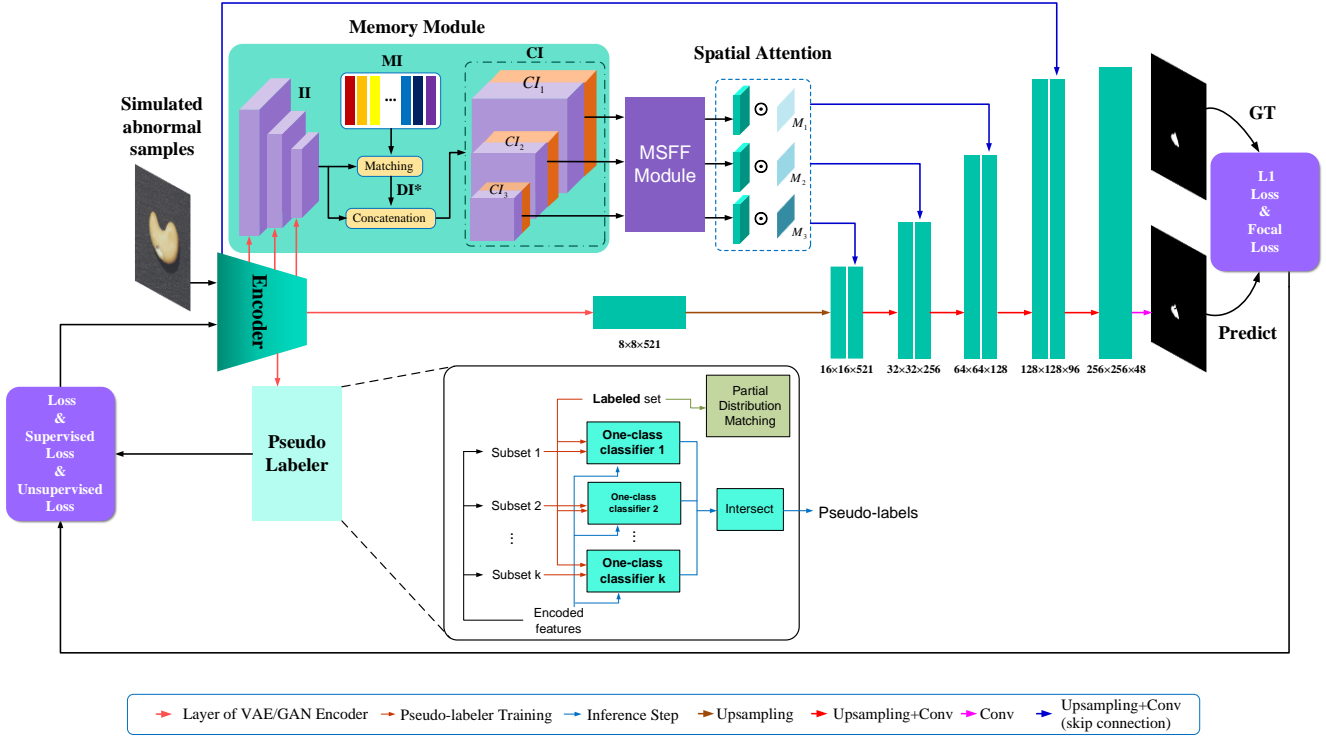

Fig. 1. Overview of the MAPL model. The model is based on the Memseg framework and incorporates an adapted VAE/GAN encoder and LeakyReLu activation function, and introduces a pseudo labeler to enhance robustness, improving the adaptability to anomaly simulation datasets and anomaly detection efficacy.

During the training phase, an anomaly simulation method is used by MAPL to incorporate artificially generated anomalous samples. These samples are created by changing the texture and structure of the normal samples $I$, resulting in the anomalous simulation dataset $\mathcal{D}^a$. The model is trained using these samples to distinguish between normal and anomalous samples. The encoder $h$ transforms the input feature $\mathbf{x}$ into the potential representation $\mathbf{r} = h(\mathbf{x})$. For this investigation, characteristics were extracted from the input image using a modified VAE/GAN-based encoder $h$. To enhance the model's ability to handle negative input values, the conventional ReLU activation function was replaced with the LeakyReLU activation function [27] in the encoder. The $\mathbf{r}$ anomaly score $q(\mathbf{r})$ is calculated by Predictor $q$ utilizing the acquired representations. The anomaly score $q(h(\mathbf{x}))$ is determined by both the encoder and the predictor.

Subsequently, pseudo labelers are introduced into the training process. The pseudo-labeler $v$, with a label set $\mathcal{H} \rightarrow \{0, 1, -1\}$, assigns pseudo-labels to the anomaly-simulated data $\mathbf{x}^u$ using OCCs. In this context, $v(h(\mathbf{x}^u)) = 1/0/-1$ indicates pseudo anomalous/pseudo normal/anomaly-simulated. With this strategy, the binary cross-entropy loss $L_{Y^u}$ is obtained for labeled data and the binary cross-entropy loss $L_{Y^l}$ is obtained for pseudo-labeled data, by which datautilization is improved and model efficiency is enhanced.

The memory module $MB$ stores the memory information $MI$ extracted from normal samples, and these features are extracted by $h$. During the inference stage, the model calculates the L2 distance between the feature $\mathbf{x}$ of the new input image and the feature in MI to obtain the difference information DI. It then finds the minimum difference value $DI^*$ between $\mathbf{x}$ and its nearest memory-like sample. Finally, it obtains the connection information $CI$, which is composed of $\mathbf{x}$ and $DI^*$.

Subsequently, the $CI$ is transferred to the Multi-scale Feature Fusion (MSFF) module, where the coordinate attention method is eliminated. This module uses a 3 × 3 convolutional layer to preserve the number of channels and modify the resolution and number of channels of feature maps with varying dimensions using up-sampling and convolutional operations. It then achieves multiscale fusion by performing element-level summation.

The spatial attention map utilizes $DI^*$ to generate three spatial attention maps, denoted as $M_n$. The information $CI$ processed through MSFF is weighted through $M_n$ and finally passed to the decoder.

Finally, the loss function module utilizes a composite loss function method, integrating L1 loss $L_{l1}$, focus loss $L_f$, and pseudo labeler-specific binary cross-entropy losses $L_{Y^u}$ and $L_{Y^l}$, which collectively optimize the performance of the model.

In order to overcome these limitations, SPADE introduces the pseudo labeler method, which utilizes consistency checking to generate pseudo labels. These labels not only streamline the model training process, but also enhance the model's ability to generalize new anomaly kinds that have not been previously seen. This combination boosts both the data use efficiency and the generalization capabilities of the model, leading to improved accuracy in anomaly identification. This is particularly beneficial in complicated industrial scenarios.

## 2.1 Abnormal Simulation

Memseg develops a highly effective self-supervised learning approach to model anomaly samples in different types of anomaly patterns found in industrial scene. This technique is executed by a series of three sequential steps:

Initially, a binary mask $M_1$ is generated by converting the input picture $I$ into a binary representation. A mask $M_P$ consisting of consecutive blocks is created through the process of threshold binarization utilizing two-dimensional Perlin noise [28]. Ultimately, a mask image $M$ is acquired by doing element-wise multiplication of the two produced masks.

The second step involves the creation of a noisy foreground picture, denoted as $I_n^{'}$, by merging the mask $M$ with the noisy image $I_n$. Additionally, a transparency factor $\delta$ is incorporated to increase the fidelity of the simulated anomalies to the actual scenario. The mathematical expression representing $I_n^{'}$ is:

$$I_n^{'} = \delta(M \odot I_n) + (1-\delta)(M \odot I) \quad (1)$$

Simultaneously, the transparency of the noisy image is increased to a higher degree, and $\delta$ is selected from the range of [0.15, 1] to intensify the difficulty of model training and boost its resilience.

The background picture is obtained by multiplying the inverse mask $\bar{M}$ with the image $I$ element by element. This background image is then added to the noisy foreground image $I_n^{'}$ to create the final simulated anomalous image $I_A$.

$$I_A = \bar{M} \odot I + I_n^{'} \quad (2)$$

Meanwhile, the noisy image $I_n$ incorporates the texture features from the DTD dataset [29] and the structural features obtained from random image alteration. This effectively creates simulated anomaly samples that encompass both texture and structural variations. During the training of MemSeg, the anomaly simulation dataset $\mathcal{D}^a$ is formed by a variable number of anomaly simulation images $I_A$. The number of images in the dataset is defined by the batch in the training batch. This method not only improves the resemblance between simulated and real anomalies, particularly in generating target foregrounds, but also creates the optimal conditions for the model to train effectively, hence improving the model's ability to detect anomalies.

## 2.2 Pseudo Labeler
### 2.2.1 Pseudo Labeler Architecture and Training

The model shown in Fig. 2 illustrates the configuration of the Pseudo labeler.

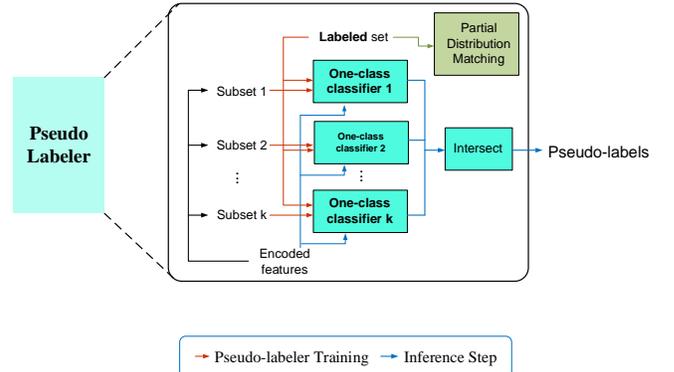

Fig. 2. Structure of Pseudo Labeler.

The pseudo-labeler comprises a model that includes $K$ $OCCs(o_1, o_2, ..., o_K)$. Each OCC is trained with negative data $(\mathcal{D}_0^l)$ and one of $K$ separate anomalous simulation subsets $(\mathcal{D}_1^a, \mathcal{D}_2^a, ..., \mathcal{D}_K^a)$. The OCC generates an anomaly score, denoted as $o_k(\mathbf{x})$, for the input sample $\mathbf{x}$. When there is consensus among all OCCs, positive pseudo-labels (i.e., predicted as anomalies) are allocated to the appropriate anomalous simulation data samples: $v(h(\mathbf{x}^a)) = 1$, if $\prod_{k=1}^{K} \hat{y}_k^{pa} = 1$, where

$$\hat{y}_k^{pa} = \begin{cases} 1 & \text{if } o_k(h(\mathbf{x}^a)) > \eta_k^p \\ 0 & \text{otherwise} \end{cases} \quad (3)$$

Likewise, the designation of negative pseudo label (i.e., normal) is assigned to a sample if all OCCs unanimously classify it as a negative pseudo label. This is based on the condition that if $\prod_{k=1}^{K} \hat{y}_k^{na} = 1$, then $v(h(\mathbf{x}^a)) = 0$, where

$$\hat{y}_k^{na} = \begin{cases} 1 & \text{if } o_k(h(\mathbf{x}^a)) < \eta_k^p \\ 0 & \text{otherwise} \end{cases} \quad (4)$$

Simulation data that deviates from the norm and lacks agreement will be classified as unknown:

$$v(h(\mathbf{x}^u)) = -1 \text{ if } \prod_{k=1}^{K} \hat{y}_k^{pa} \times \hat{y}_k^{na} = 0$$

### 2.2.1 Regarding the selection of thresholds $\eta^p$ and $\eta^n$

SPADE [30] utilizes the partial matching method to estimate the marginal distribution of anomalous simulated data. This method matches the distribution to a known single class (positive or negative) distribution without compromising the validation set including labeled data. The general principle states that normal samples are close to each other in feature space, while abnormal samples likewise tend to be close to each other in features. This study aims to identify the marginal distribution of positively labeled samples by comparing their distribution of abnormal scores to the distribution of abnormal scores of artificially generated abnormal samples. The value of $\eta^p$ is then determined based on this comparison. The identical approach is employed to ascertain $\eta^n$ by utilizing negatively labeled data. $\eta^p$ and $\eta^n$ are defined according to equations (5) and (6).

$$\eta_k^p = \arg\min_{\eta} D_w(\{o_k(h(\mathbf{x}^l) \mid y^l = 1\}, \{o_k(h(\mathbf{x}^a)) > \eta\}) \quad (5)$$

$$\eta_k^n = \arg\min_{\eta} D_w(\{o_k(h(\mathbf{x}^l) \mid y^l = 0\}, \{o_k(h(\mathbf{x}^a)) < \eta\}) \quad (6)$$

where $D_w$ is the Wasserstein distance between the two distributions on which the determination of the subset of anomalous simulated data for pseudo-labeling is based.

## 2.3 Memory Module

The MemSeg model incorporates a memory module that is influenced by the human process of detecting irregularities. It achieves this by utilizing knowledge of the normal state and comparing the observed image with the stored normal image to find abnormal regions. In order to emulate this process, a collection of specifically chosen normal images is employed as memory samples and high-level characteristics are extracted from these samples by the encoder $h$.

Our $h$ employs a multilayer convolutional network structure to capture image features ranging from coarse to fine. Specifically, features obtained from layers 2, 3, and 4 are utilized, which provide feature maps with dimensions $N \times 64 \times 64 \times 64$, $N \times 128 \times 32 \times 32$, and $N \times 256 \times 16 \times 16$, respectively. Collectively, these distinct resolutions of features form what is known as memory information ($MI$), which supplies a wealth of visual data for detecting anomalies in the future.

$h$ extracts dimensions of $64 \times 64 \times 64$, $128 \times 32 \times 32$, and $256 \times 16 \times 16$ from the input image during the training or inference phase. The information ($\mathbf{x}$) of the input image is comprised of these multi-scale features. To calculate the L2 distance between the input image and memory sample, $N$ difference information $DI$ is determined using the following:

$$DI = \bigcup_{i=1}^{N} \| MI_i - \mathbf{x} \|_2 \quad (7)$$

Where $N$ is the number of memory samples. $DI$ is used to choose the most similar memory feature for each input characteristic in order to identify discrepancies. The minimum L2 distance signifies the existence of possibly abnormal regions, with greater values indicating a greater probability of anomalies. Additionally, it undergoes further filtration to determine the optimal difference information $DI^*$, a variable that represents the minimum difference value between $\mathbf{x}$ and its closest memory sample:

$$DI^* = \underset{DI_i \in DI}{\arg\min} \sum_{x \in DI_i} x \quad (8)$$

The $DI^*$ metric quantifies the extent to which the input image differs from the most similar memory samples. A higher value suggests a greater likelihood of an anomaly occurring at the corresponding region. The $DI^*$ and $\mathbf{x}$ are combined in the channel dimension to create a spliced feature, resulting in the concatenated information $CI_1$, $CI_2$, and $CI_3$. The multi-scale feature fusion module processes this feature and then transfers it to the decoder via the U-Net jump link to identify anomalous regions. The multi-scale feature fusion(MSFF) module processes this feature and then it is transmitted to the decoder $g$ for the identification of anomalous regions through the jump connection of U-Net.

## 2.4 Spatial Attention Map

The approach creates three spatial attention maps—images that aid in more precisely predicting important disparities in anomalous regions—to maximize the utilization of disparity information. The average value of the features of the various $DI^*$ dimensions in the channel dimension is computed to create three feature maps of varying sizes (16 x 16, 32 x 32, and 64 x 64). Within this method, the $M_3$ spatial attention map is directly derived from the 16 x 16 feature map. $M_2$ is then obtained by up-sampling and multiplying $M_3$ at the element level by the $32 \times 32$ feature map. Ultimately, $M_2$ undergoes additional up-sampling and element-level operations are coupled with the $64 \times 64$ feature map to create $M_1$. The spatial attention maps $M_1$, $M_2$, and $M_3$ were individually weighted for information $CI_1$, $CI_2$, and $CI_3$ via MSFF processing, as seen in Fig. 1. The precise equation is as follows:

$$M_3 = \frac{1}{C_3}\sum_{i=1}^{C_3}DI^*_{3i} \quad (9)$$

$$M_2 = \frac{1}{C_2}\left(\sum_{i=1}^{C_2}DI^*_{2i}\right)\odot M_3^U \quad (10)$$

$$M_1 = \frac{1}{C_1}\left(\sum_{i=1}^{C_1}DI^*_{1i}\right)\odot M_2^U \quad (11)$$

Where $C_n$ represents the number of $DI_n^*$ channels, $DI^*_{ni}$ represents the feature mapping of channel $i$ in $DI_n^*$. $M_n^U$ represents the feature map obtained after up-sampling $M_n$.

## 2.5 Multi-Scale Feature Fusion Module

The utilization of $CI$ directly is hindered by feature redundancy, leading to an increased computational load on the model and thus impacting the pace of inference. Building upon the success of multiscale feature fusion in target identification [31], [32], the model utilizes a mix of a channel attention mechanism and a multiscale feature fusion method to enhance the visual and semantic information in $CI$.

The fusion module initiates the fusion process by employing a 3x3 convolutional layer to maintain a consistent number of channels. Despite the initial belief that the incorporation of Coordinate Attention (CA) [33] would strengthen the connection between different features, studies conducted in Section 3.5 demonstrated that it did not lead to any meaningful performance enhancement. Thus, in order to enhance efficiency, the decision was made to eliminate this method in order to decrease the computational load without sacrificing performance.

Afterwards, the feature maps of varying dimensions are aligned in terms of resolution and number of channels using up-sampling and convolution operations. Then, multi-scale feature fusion is achieved by summing the elements at each level.

Ultimately, the process of combining multi-scale features is accomplished through the inclusion of elements at a granular level. The combined feature maps are multiplied by the spatial attention map $M_n$ and then fed into the final decoder to enhance the effectiveness of the model in decoding intricate situations.

## 2.6 Loss Function and Optimization

### 2.6.1 Composite Loss Function Design and Application

In this study, L1 loss and focal loss [34] are utilized in the original Memseg model to guarantee the uniformity of all pixels in the image space. The L1 loss maintains the image edge details, but the focal loss balances the area weight between normal and aberrant parts in the picture, hence enhancing the model's ability to distinguish challenging samples that are difficult to segment. The formula is as stated:

$$L_{l1} = \| G - \hat{G}\|_1 \quad (12)$$

$$L_f = -\alpha_t(1-p_t)^\gamma log(p_t) \quad (13)$$

where $G$ is the ground truth of the anomalous region in the anomalous simulation dataset, and $\hat{G}$ is the predicted value. $p_t$ denotes the predicted probability $p$ when the corresponding pixel has a true label of 1 in $G$, and similarly $p_t = 1-p$ if the true label of $G$ is 0. Meanwhile, the two hyperparameters, $\gamma$ and $\alpha_t$, are utilized to fine-tune the intensity of the weight application.

Ultimately, these limitations are combined to form an objective function:

$$L_m = \omega_{l1}L_{l1} + \omega_f L_f \tag{14}$$

where $\omega_{l1}$ and $\omega_f$ are the weighting factors for the different loss functions. Initially, in the training phase, a fundamental composite objective function (Eqs. 14) is established. This function is adjusted to cater to the diverse optimization requirements of the model by merging the L1 loss and the focus loss. Furthermore, in order to improve the model's capacity to handle abnormal simulated data, the loss function of the pseudo labeler is integrated into the overall objective function.

### 2.6.2 Pseudo Labeler Loss Function Strategy

In the training phase of pseudo-labeling, two types of loss functions are utilized: a supervised binary cross entropy (BCE) loss [35] for labeled data, and an unsupervised BCE loss for tackling pseudo-labeled data. This combination enhances the overall efficacy of the model in both supervised and unsupervised learning situations. More precisely, the supervised loss is determined by the binary cross entropy (BCE) loss function, which is used on labeled data to guarantee that the model's predictions align with the true labels. The purpose of this loss function is to incentivize the model to effectively differentiate between normal and abnormal samples.

$$L_{Y^l} = \mathbb{E}[L_{BCE}(q(h(\mathbf{x}^l)), y^l)] \tag{15}$$

where $q(h(\mathbf{x}^l)$ denotes the $\mathbf{x}^l$ prediction result of the model for labeled data and $y^l$ is the corresponding true label.

The unsupervised loss is utilized on pseudo-labeled data, and the automatically generated labels are employed for training the model in order to enhance its ability to generalize to unfamiliar data.

$$L_{Y^u} = \mathbb{E}[L_{BCE}(q(h(\mathbf{x}^u)), v(h(\mathbf{x}^u))) \times 1\{v^u \in \{0,1\}\}] \tag{16}$$

where $v(h(\mathbf{x}^u))$ represents the pseudo-labeled prediction results for the anomalous simulated data $\mathbf{x}^u$, generated by Pseudo Labeler. $1\{v^u \in \{0,1\}\}$ is a binary weight that adaptively modifies the influence of the loss incurred from pseudo-labeled data based on the present performance of the model.

The ultimate objective of the model's training is to minimize the combined losses of these two factors along with the Memseg loss, which can be mathematically represented as:

$$L_{all} = L_m + L_{Y^u} + L_{Y^l} \tag{17}$$

### 2.6.3 Training optimization and convergence monitoring

During model training, the combined loss function $L_{all}$ is applied, designed to optimize encoder ($h$), predictor ($q$), Memory bank ($MB$), MSFF ($MF$) and Decoder ($g$).

$$h^*, q^*, MB^*, MF^*, g^* = \arg\min_{h,q,MB,MF,d} L_{all} \tag{18}$$

Throughout the training phase, the fluctuation in loss is carefully observed. If the drop in loss does not exhibit a significant decline during multiple consecutive training iterations, it might be inferred that the model is nearing convergence. At this juncture, the option is available to either end the training process or modify parameters, such as the learning rate, in order to enhance the performance of the model.

## 3 Experiment

### 3.1 Dataset

To accurately assess the effectiveness and feasibility of the proposed MAPL method, this study developed a new dataset called BHAD (Benchmark for High precision Anomaly Detection)[36]. The dataset has been made public on the Mendeley Data platform to support comprehensive validation of the MAPL method. The main objective of this dataset is to give a full evaluation of the MAPL Net approach. This dataset integrates specific components from established datasets such as MVTec AD, Visa, and MDPP. It undergoes preprocessing and calibration for each dataset type to ensure consistent training circumstances and data coherence. The BHAD dataset covers four different object types, each providing 369 to 432 normal training samples and test images containing normal and abnormal samples. The test set is specifically designed to include authentic faults with diverse textures and sizes, in order to thoroughly evaluate the model's capacity to detect them.

An exception mask configuration is established by taking into account the dataset's characteristics, such as mask utilization, adjustment of background threshold, and background inversion. This configuration aims to enhance the preprocessing of the data. In order to replicate interference in real-world industrial scenarios, random Gaussian noise and contrast modification were introduced to the images in the dataset. It is designed to improve the model's ability to handle environmental interference in actual settings. The performance evaluation utilizes image-level AUC-ROC metrics to assure precision and dependability.

### 3.2 Experimental Methods

For an exhaustive demonstration of the experimental methodology, a pseudo-code for the MAPL model (Algorithm 1) is provided, which describes in detail the key steps from data preprocessing to model training. This includes the generation of pseudo-labels and the processing of anomalous simulated data, guaranteeing the reproducibility of the experiments.

---

**Algorithm 1** Memory Augmentation and Pseudo-Labeling for Semi-Supervised Anomaly Detection (MAPL)

**Input**: Labeled training data $\mathcal{D}^l$

**Output**: Trained encoder ($h$), predictor ($q$), Memory bank ($MB$), MSFF($MF$), Decoder($g$)

---

1: **function** PSEUDO-LABELER($\mathcal{D}_1^l, \mathcal{D}_0^l, \mathcal{D}^a, h$)
2:    Split the dataset $\mathcal{D}^a$ into $K$ non-overlapping subsets $\{\mathcal{D}_k^a\}_{k=1}^K$
3:    **for** $k=1:K$ **do**
4:      Train OCC models $o_k$ on $\mathcal{D}_k^a \cup \mathcal{D}_0^l$
5:      Assign $\eta_k^p / \eta_k^n$ by applying partial matching to $\mathcal{D}_1^l, \mathcal{D}_0^l$ following Eqs. 5 and 6.
6:    **end for**
7:    Build pseudo-labeler $v$ using Eqs. 3 and 4.
8:    **Return** pseudo-labeler $v$.
9: end function
10: **function** Train_MAPL($h, q, MB, MF, g, \mathcal{D}^l$)
11:    **for** each step **do**
12:      **for** each batch from $\mathcal{D}^l$ **do**
13:        Extract samples from $\mathcal{D}^l$ and generate anomaly simulation samples to form an anomaly simulation dataset $\mathcal{D}^a$ following Eqs. 1 and 2.
14:        **Set** positively / negatively labeled data $\mathcal{D}_1^l, \mathcal{D}_0^l$
15:        $v$=PSEUDO-LABELER($\mathcal{D}_1^l, \mathcal{D}_0^l, \mathcal{D}^a, h$)
16:        Merge $DI^*$ and $\mathcal{D}^l$ in the channel dimension to obtain concatenated information $CI$
17:        Extract three spatial attention maps $M$ using $DI^*$ following Eqs. 9, 10, and 11.
18:        Using $g$ to predict using $M$ to weight the $CI$ processed by $MF$
19:        Calculate $L_m$ based on L1 and focal loss using Eqs. 15, 16, and 17.
20:      end for
21:    end for
22:    **Return** $h, q, MB, MF, g$
23: end function
24: **Initialize** $h, q, MB, MF, g$
25: Extract features from $N$ samples in $\mathcal{D}^l$ using $h$ as memory samples to form a memory information $MI$ and update $MB$
26: **while** $h, q, MB, MF, g$ not converged **do**
27:    $h, q, MB, MF, g$ =Train_MAPL($h, q, MB, MF, g, \mathcal{D}^l, \mathcal{D}^t$)
28:    Update $h, q, MB, MF, g$ using Eq. 18.
29: end while

### 3.3 Experimental Environment and Hyperparameter Settings

The experimental environment configuration is shown in Table 1:

| Name | Parameter |
| --- | --- |
| Operating System | Ubuntu22.04 |
| CPU | Xeon(R) Gold 6430 CPU |
| RAM | 120 GB |
| GPU | NVIDIA RTX 4090 (24 GB) |
| CUDA | 12.1 |
| Deep learning framework | PyTorch 2.1.0 |

Table 1: Detailed configuration parameters of experimental environment

The following strategies, as shown in Table 2, are adopted to ensure the efficient training of the MAPL model:

| Hyper-parameters | Parameter |
|---|---|
| Iteration steps | 500 |
| Batch_size | 8 (4 normal samples and 4 simulated abnormal samples) |
| Image_size | 256×256 |
| Learning rate | 0.003 |
| Focus loss | 4 |
| Loss function weighting factor | 0.6 |
| Loss function weighting factor | 0.4 |
| Number of memory samples | 30 |
| OCC settings | Gaussian Mixture Model (GMM) |
| Number of OCC | 2 |

Table 2: Parameter configuration for model training

In this case, the learning rate is determined by grid search and is initially set to 0.003, followed by an initial adjustment using a warm up strategy, and finally the cosine annealing algorithm is applied for dynamic decay. Meanwhile, in order to balance the model's ability to handle different anomaly types, texture and structure anomaly simulations are set to be performed with equal probability. The number of training rounds of the pseudo labeler is dynamically adjusted according to the size of the dataset to ensure that the GMM can be effectively updated in each round of data iteration.

### 3.4 Comparative Experiment

On the BHAD dataset, MAPL outperforms the original MemSeg model in terms of image-level ROC-AUC scores. As shown in Table 3, MAPL achieves higher performance in all subcategories, especially in the Cashew and Leather categories.

| Datasets | Bracket black | Cashew | Leather | Screw | Average |
|---|---|---|---|---|---|
| Memseg | 75.23 | 78.00 | 96.71 | 74.46 | 81.10 |
| MAPL（Ours） | **81.07** | **87.17** | **98.78** | **76.67** | **86.20** |

Table 3: Image-level ROC-AUC % performance comparison between MAPL and MemSeg on the BHAD dataset

MAPL shows significant advantages in processing complex and texture-rich industrial images, which validates its effectiveness in enhancing model robustness and accuracy, especially in the presence of noise interference and complex textures.

### 3.5 Ablation Experiments

Ablation experiments play an imperative role in evaluating the impact of various model components on overall performance. This study specifically focuses on whether or not to remove the effects of CA, spatial attention maps, and MSFF. The following are the primary experiments conducted:

| Datasets | Bracket black | Cashew | Leather | Screw | Average |
|---|---|---|---|---|---|
| No msff | 75.23 | 78.00 | 96.71 | 74.46 | 82.28 |
| Added CA | **81.07** | **87.17** | 98.78 | 76.67 | 85.92 |
| No Spatial Attention | 46.04 | 78.89 | **99.35** | 71.12 | 73.85 |
| MAPL（Ours） | 78.01 | 85.62 | 99.21 | **81.96** | **86.20** |

Table 4: Image-level ROC-AUC % performance comparison of MAPL's ablation experiments on the BHAD dataset

The data demonstrates that removing MSFF has a substantial impact on the model's performance, resulting in a decrease in the average AUROC to 82.28%. This suggests that MSFF is imperative for feature fusion and minimizing information redundancy. Although CA enhances performance in some configurations, the overall MAPL model without CA applied demonstrates the highest average performance across all test settings. Eliminating the spatial attention maps led to a notable decline in performance on certain datasets, highlighting its significance in identifying abnormal locations.

Based on the results of the ablation experiments, the exclusion of CA and the retention of MSFF and spatial attention maps were decided upon due to their demonstrated significant effectiveness. This strategic config-uration, by concentrating on the most influential components, has led to a simplified model framework and a reduction in computational burden, without substantially affecting performance, as evidenced by the highest observed average AUROC scores. Through such a configuration strategy, the efficiency of computational resource utilization has been optimized, and the robustness and accuracy of the model in various anomaly detection scenarios have been ensured. Furthermore, the key to configuring the modules involves a systematic assessment of each component's contribution to the overall model performance, followed by selective application or exclusion to achieve an optimal balance between performance and resource utilization. According to the above experiments, it can be obtained that the MAPL model performs optimally under all test conditions, with an average AUROC of 86.20%, which is 5.1% higher than MemSeg's 81.10%, and fully proves the synergistic effect between the model components.

## 4 Conclusion

The paper proposes the MAPL model, which obtains an average image-level AUROC score of 86.2% on the BHAD dataset. This model significantly improves the accuracy and resilience of industrial surface defect identification, particularly in challenging conditions with noise interference. This outcome not only demonstrates exceptional efficacy in handling intricate situations with noise interference but also presents novel methodologies for study and practical applications in the realm of industrial anomaly detection. In the real world, obtaining high-quality anomaly data is a challenge, and this challenge highlights the need to further optimize and validate the MAPL model for application in different real-world environments. Therefore, future work will concentrate on expanding the application scenarios and improving the model by incorporating additional practical environments for further validation and optimization of the MAPL model. These explorations will tackle the current restrictions and improve the model's capacity to generalize and be practically applied.

In summary, this research tackles the challenges of handling unlabeled large-scale data and detecting difficult-to-identify anomalies in industrial settings.

## References

[1] P. Bergmann, M. Fauser, D. Sattlegger, and C. Steger, "MVTec AD -- A Comprehensive Real-World Dataset for Unsupervised Anomaly Detection," presented at the Proceedings of the IEEE/CVF Conference on Computer Vision and Pattern Recognition, 2019, pp. 9592–9600. Accessed: Feb. 15, 2024. [Online]. Available: https://openaccess.thecvf.com/content_CVPR_2019/html/Bergmann_MVTec_AD_--_A_Comprehensive_Real-World_Dataset_for_Unsupervised_Anomaly_CVPR_2019_paper.html

[2] Y. Zou, J. Jeong, L. Pemula, D. Zhang, and O. Dabeer, "SPot-the-Difference Self-supervised Pre-training for Anomaly Detection and Segmentation," in *Computer Vision – ECCV 2022*, Springer, Cham, 2022, pp. 392–408. doi: 10.1007/978-3-031-20056-4_23.

[3] S. Jezek, M. Jonak, R. Burget, P. Dvorak, and M. Skotak, "Deep learning-based defect detection of metal parts: evaluating current methods in complex conditions," in *2021 13th International Congress on Ultra Modern Telecommunications and Control Systems and Workshops (ICUMT)*, Oct. 2021, pp. 66–71. doi: 10.1109/ICUMT54235.2021.9631567.

[4] M. Yang, P. Wu, and H. Feng, "MemSeg: A semi-supervised method for image surface defect detection using differences and commonalities," *Eng. Appl. Artif. Intell.*, vol. 119, p. 105835, Mar. 2023, doi: 10.1016/j.engappai.2023.105835.

[5] P. Bergmann, S. Löwe, M. Fauser, D. Sattlegger, and C. Steger, "Improving Unsupervised Defect Segmentation by Applying Structural Similarity to Autoencoders," in *Proceedings of the 14th International Joint Conference on Computer Vision, Imaging and Computer Graphics Theory and Applications*, 2019, pp. 372–380. doi: 10.5220/0007364503720380.

[6] T.-W. Tang, W.-H. Kuo, J.-H. Lan, C.-F. Ding, H. Hsu, and H.-T. Young, "Anomaly Detection Neural Network with Dual Auto-Encoders GAN and Its Industrial Inspection Applications," *Sensors*, vol. 20, no. 12, Art. no. 12, Jan. 2020, doi: 10.3390/s20123336.

[7] D. T. Nguyen, Z. Lou, M. Klar, and T. Brox, "Anomaly Detection With Multiple-Hypotheses Predictions," in *Proceedings of the 36th International Conference on Machine Learning*, PMLR, May 2019, pp. 4800–4809. Accessed: Mar. 20, 2024. [Online]. Available: https://proceedings.mlr.press/v97/nguyen19b.html

[8] M. Sakurada and T. Yairi, "Anomaly Detection Using Autoencoders with Nonlinear Dimensionality Reduction," in *Proceedings of the MLSDA 2014 2nd*

*Workshop on Machine Learning for Sensory Data Analysis*, in MLSDA'14. New York, NY, USA: Association for Computing Machinery, Dec. 2014, pp. 4–11. doi: 10.1145/2689746.2689747.

[9] T. Schlegl, P. Seeböck, S. M. Waldstein, U. Schmidt-Erfurth, and G. Langs, "Unsupervised Anomaly Detection with Generative Adversarial Networks to Guide Marker Discovery," in *Information Processing in Medical Imaging*, M. Niethammer, M. Styner, S. Aylward, H. Zhu, I. Oguz, P.-T. Yap, and D. Shen, Eds., in Lecture Notes in Computer Science. Cham: Springer International Publishing, 2017, pp. 146–157. doi: 10.1007/978-3-319-59050-9_12.

[10] S. Akcay, A. Atapour-Abarghouei, and T. P. Breckon, "GANomaly: Semi-supervised Anomaly Detection via Adversarial Training," in *Computer Vision – ACCV 2018*, C. V. Jawahar, H. Li, G. Mori, and K. Schindler, Eds., in Lecture Notes in Computer Science. Cham: Springer International Publishing, 2019, pp. 622–637. doi: 10.1007/978-3-030-20893-6_39.

[11] H. Zenati, C. S. Foo, B. Lecouat, G. Manek, and V. R. Chandrasekhar, "Efficient GAN-Based Anomaly Detection." arXiv, May 01, 2019. doi: 10.48550/arXiv.1802.06222.

[12] L. Bergman, N. Cohen, and Y. Hoshen, "Deep Nearest Neighbor Anomaly Detection," arXiv.org. Accessed: Mar. 20, 2024. [Online]. Available: https://arxiv.dosf.top/abs/2002.10445v1

[13] N. Cohen and Y. Hoshen, "Sub-Image Anomaly Detection with Deep Pyramid Correspondences," arXiv.org. Accessed: Mar. 20, 2024. [Online]. Available: https://arxiv.dosf.top/abs/2005.02357v3

[14] Y. Zheng, X. Wang, R. Deng, T. Bao, R. Zhao, and L. Wu, "Focus Your Distribution: Coarse-to-Fine Non-Contrastive Learning for Anomaly Detection and Localization," in *2022 IEEE International Conference on Multimedia and Expo (ICME)*, Jul. 2022, pp. 1–6. doi: 10.1109/ICME52920.2022.9859925.

[15] K. Roth, L. Pemula, J. Zepeda, B. Schölkopf, T. Brox, and P. Gehler, "Towards Total Recall in Industrial Anomaly Detection," presented at the Proceedings of the IEEE/CVF Conference on Computer Vision and Pattern Recognition, 2022, pp. 14318–14328. Accessed: Mar. 21, 2024. [Online]. Available: https://openaccess.thecvf.com/content/CVPR2022/html/Roth_Towards_Total_Recall_in_Industrial_Anomaly_Detection_CVPR_2022_paper.html

[16] T. Defard, A. Setkov, A. Loesch, and R. Audigier, "PaDiM: A Patch Distribution Modeling Framework for Anomaly Detection and Localization," in *Pattern Recognition. ICPR International Workshops and Challenges*, Springer, Cham, 2021, pp. 475–489. doi: 10.1007/978-3-030-68799-1_35.

[17] M. Salehi, N. Sadjadi, S. Baselizadeh, M. H. Rohban, and H. R. Rabiee, "Multiresolution Knowledge Distillation for Anomaly Detection," presented at the Proceedings of the IEEE/CVF Conference on Computer Vision and Pattern Recognition, 2021, pp. 14902–14912. Accessed: Mar. 21, 2024. [Online]. Available: https://openaccess.thecvf.com/content/CVPR2021/html/Salehi_Multiresolution_Knowledge_Distillation_for_Anomaly_Detection_CVPR_2021_paper.html

[18] G. Wang, S. Han, E. Ding, and D. Huang, "Student-Teacher Feature Pyramid Matching for Anomaly Detection," arXiv.org. Accessed: Mar. 21, 2024. [Online]. Available: https://arxiv.dosf.top/abs/2103.04257v3

[19] T. D. Tien *et al.*, "Revisiting Reverse Distillation for Anomaly Detection," presented at the Proceedings of the IEEE/CVF Conference on Computer Vision and Pattern Recognition, 2023, pp. 24511–24520. Accessed: Mar. 21, 2024. [Online]. Available: https://openaccess.thecvf.com/content/CVPR2023/html/Tien_Revisiting_Reverse_Distillation_for_Anomaly_Detection_CVPR_2023_paper.html

[20] P. Bergmann, M. Fauser, D. Sattlegger, and C. Steger, "Uninformed Students: Student-Teacher Anomaly Detection With Discriminative Latent Embeddings," presented at the Proceedings of the IEEE/CVF Conference on Computer Vision and Pattern Recognition, 2020, pp. 4183–4192. Accessed: Mar. 21, 2024. [Online]. Available: https://openaccess.thecvf.com/content_CVPR_2020/html/Bergmann_Uninformed_Students_Student-Teacher_Anomaly_Detection_With_Discriminative_Latent_Embeddings_CVPR_2020_paper.html

[21] H. Deng and X. Li, "Anomaly Detection via Reverse Distillation From One-Class Embedding," presented at the Proceedings of the IEEE/CVF Conference on Computer Vision and Pattern Recognition, 2022, pp. 9737–9746. Accessed: Mar. 21, 2024. [Online]. Available: https://openaccess.thecvf.com/content/CVPR2022/html/Deng_Anomaly_Detection_via_Reverse_Distillation_From_One-Class_Embedding_CVPR_2022_paper.html

[22] T. Cao, J. Zhu, and G. Pang, "Anomaly Detection Under Distribution Shift," presented at the Proceedings of the IEEE/CVF International Conference on Computer Vision, 2023, pp. 6511–6523. Accessed: Mar. 21,

2024. [Online]. Available: https://openaccess.thecvf.com/content/ICCV2023/html/Cao_Anomaly_Detection_Under_Distribution_Shift_ICCV_2023_paper.html

[23] X. Zhang, S. Li, X. Li, P. Huang, J. Shan, and T. Chen, "DeSTSeg: Segmentation Guided Denoising Student-Teacher for Anomaly Detection," presented at the Proceedings of the IEEE/CVF Conference on Computer Vision and Pattern Recognition, 2023, pp. 3914–3923. Accessed: Mar. 21, 2024. [Online]. Available: https://openaccess.thecvf.com/content/CVPR2023/html/Zhang_DeSTSeg_Segmentation_Guided_Denoising_Student-Teacher_for_Anomaly_Detection_CVPR_2023_paper.html

[24] A. B. L. Larsen, S. K. Sønderby, H. Larochelle, and O. Winther, "Autoencoding beyond pixels using a learned similarity metric," in *Proceedings of The 33rd International Conference on Machine Learning*, PMLR, Jun. 2016, pp. 1558–1566. Accessed: Feb. 15, 2024. [Online]. Available: https://proceedings.mlr.press/v48/larsen16.html

[25] J. Yoon, K. Sohn, C.-L. Li, S. O. Arik, and T. Pfister, "SPADE: Semi-supervised Anomaly Detection under Distribution Mismatch." arXiv, Nov. 30, 2022. doi: 10.48550/arXiv.2212.00173.

[26] O. Ronneberger, P. Fischer, and T. Brox, "U-Net: Convolutional Networks for Biomedical Image Segmentation," in *Medical Image Computing and Computer-Assisted Intervention – MICCAI 2015*, Springer, Cham, 2015, pp. 234–241. doi: 10.1007/978-3-319-24574-4_28.

[27] A. L. Maas, "Rectifier Nonlinearities Improve Neural Network Acoustic Models," 2013. Accessed: Mar. 07, 2024. [Online]. Available: https://www.semanticscholar.org/paper/Rectifier-Nonlinearities-Improve-Neural-Network-Maas/367f2c63a6f6a10b3b64b8729d601e69337ee3cc

[28] K. Perlin, "An image synthesizer," *ACM SIGGRAPH Comput. Graph.*, vol. 19, no. 3, pp. 287–296, Jul. 1985, doi: 10.1145/325165.325247.

[29] M. Cimpoi, S. Maji, I. Kokkinos, S. Mohamed, and A. Vedaldi, "Describing Textures in the Wild," presented at the Proceedings of the IEEE Conference on Computer Vision and Pattern Recognition, 2014, pp. 3606–3613. Accessed: Apr. 22, 2024. [Online]. Available: https://openaccess.thecvf.com/content_cvpr_2014/html/Cimpoi_Describing_Textures_in_2014_CVPR_paper.html

[30] M. Christoffel, G. Niu, and M. Sugiyama, "Class-prior Estimation for Learning from Positive and Unlabeled Data," in *Asian Conference on Machine Learning*, PMLR, Feb. 2016, pp. 221–236. Accessed: Apr. 19, 2024. [Online]. Available: https://proceedings.mlr.press/v45/Christoffel15.html

[31] S. Chen, Z. Cheng, L. Zhang, and Y. Zheng, "SnipeDet: Attention-guided pyramidal prediction kernels for generic object detection," *Pattern Recognit. Lett.*, vol. 152, pp. 302–310, Dec. 2021, doi: 10.1016/j.patrec.2021.10.026.

[32] T.-Y. Lin, P. Dollar, R. Girshick, K. He, B. Hariharan, and S. Belongie, "Feature Pyramid Networks for Object Detection," presented at the Proceedings of the IEEE Conference on Computer Vision and Pattern Recognition, 2017, pp. 2117–2125. Accessed: Apr. 24, 2024. [Online]. Available: https://openaccess.thecvf.com/content_cvpr_2017/html/Lin_Feature_Pyramid_Networks_CVPR_2017_paper.html

[33] Q. Hou, D. Zhou, and J. Feng, "Coordinate Attention for Efficient Mobile Network Design," presented at the Proceedings of the IEEE/CVF Conference on Computer Vision and Pattern Recognition, 2021, pp. 13713–13722. Accessed: Apr. 24, 2024. [Online]. Available: https://openaccess.thecvf.com/content/CVPR2021/html/Hou_Coordinate_Attention_for_Efficient_Mobile_Network_Design_CVPR_2021_paper.html

[34] T.-Y. Lin, P. Goyal, R. Girshick, K. He, and P. Dollar, "Focal Loss for Dense Object Detection," presented at the Proceedings of the IEEE International Conference on Computer Vision, 2017, pp. 2980–2988. Accessed: Apr. 24, 2024. [Online]. Available: https://openaccess.thecvf.com/content_iccv_2017/html/Lin_Focal_Loss_for_ICCV_2017_paper.html

[35] U. Ruby and V. Yendapalli, "Binary cross entropy with deep learning technique for Image classification," *Int. J. Adv. Trends Comput. Sci. Eng.*, vol. 9, Oct. 2020, doi: 10.30534/ijatcse/2020/175942020.

[36] J. Chen, "Benchmark for High precision Anomaly Detection," vol. 1, May 2024, doi: 10.17632/957mmypbjy.1.